%% file: paper.tex
\documentclass{article} 
\usepackage{nips15submit_e,times}
\usepackage{graphicx}
\usepackage{amsmath,amssymb} 
\usepackage{multirow}
\usepackage{amsthm}
\usepackage{forloop}
\usepackage{calc}
\usepackage{rotating}
\usepackage{enumitem}
\usepackage{xspace}
\usepackage{pgfplots}
\usepackage{tikz}
\usepackage{wrapfig}

\usepackage[T1]{fontenc}

\title{Efficient Non-greedy Optimization of Decision Trees}
\author{
\hspace{-.45cm} Mohammad Norouzi$^{1}$\thanks{Part of this work was
  done while M. Norouzi and M. D. Collins were at Microsoft
  Research, Cambridge.} \hspace*{1cm} Maxwell
D. Collins$^{2\,*}$ \hspace*{1cm} Matthew Johnson$^3$\\ {\bf David
  J.\ Fleet$^4$} \hspace*{1cm} {\bf Pushmeet Kohli$^5$}\\[.1cm]
$^{1,4}$\:Department of Computer Science, University of
Toronto\\ $^{2}$\:Department of Computer Science, University of
Wisconsin-Madison\\ $^{3,5}\:$Microsoft Research
}

\input defs.tex

\nipsfinalcopy 

\begin{document}

\maketitle

\begin{abstract}
\input{abs}
\end{abstract}

\input{intro}
\input{previous}
\input{formulation}
\input{learning}
\input{opt}
\input{expts}
\input{discussion.tex}

\small{ \bibliography{myrefs} }
\bibliographystyle{plain}

\pagebreak
\appendix
\input{appendix.tex}

\end{document}

%% file: defs.tex
\usepackage{bm}
\usepackage{dsfont}
\usepackage{amsfonts}
\usepackage{amsmath}
\usepackage{relsize}
\usepackage{algorithmic,algorithm}
\usepackage{bm}
\usepackage{tikz}

\newcommand{\p}{p}
\newcommand{\x}{\vec{x}}
\newcommand{\y}{{y}}

\newcommand{\W}{W}
\newcommand{\w}{\vec{w}}
\newcommand{\loss}{\ell}

\newcommand{\m}{m}
\newcommand{\h}{\vec{h}}
\newcommand{\g}{\vec{g}}
\newcommand{\any}{\,.\,}

\def\by[#1]#2{#1~\hspace*{-.15cm}~\times~\hspace*{-.15cm}~#2}

\newcommand{\depth}{d}

\newcommand{\test}{s}
\newcommand{\leaf}{\vec{\theta}}
\newcommand{\Leaves}{\Theta}
\newcommand{\Hset}{\mathcal H^\m}
\newcommand{\Bset}{\mathcal B}

\newcommand{\Dset}{\mathcal D}

\newcommand{\calL}{\mathcal L}

\renewcommand{\vec}[1]{\operatorname{vec}}
\let\originaleqref\eqref
\renewcommand{\eqref}{\originaleqref}

\renewcommand{\th}[1]{#1^\text{th}}

\newcommand{\trans}[1]{{#1}^{\ensuremath{\mathsf{T}}}}

\newcommand{\slfrac}[2]{\left.#1\middle/#2\right.}

\def\vec#1{{\boldsymbol{\mathbf #1}}}

\def\Real{\mathbb{R}}
\newcommand{\Bin}[1]{\mathbb H^{#1}}
\newcommand{\Ind}[1]{\mathbb I_{#1}}

\newcommand{\f}[1]{f}

\newcommand{\figref}[1]{Fig.~\ref{#1}}
\newcommand{\propref}[1]{Proposition~\ref{#1}}

\newcommand{\sign}[0]{\operatorname{sgn}\xspace}
\newcommand{\argmax}[1]{\underset{#1}{\operatorname{argmax}}}

\newcommand{\itert}[1]{{#1}^{(t)}}
\newcommand{\itertpp}[1]{{#1}^{(t+1)}}
\newcommand{\itertph}[1]{{#1}^{(tmp)}}
\newcommand{\iterz}[1]{{#1}^{(0)}}
\newcommand{\sumxy}[0]{\hspace{-.1cm}\sum_{(\x,\y) \in \Dset}\hspace{-.1cm}}

\newcommand{\1}[1]{\mathds{1}}

\def\ie{{\em i.e.,}}

\def\eg{{\em e.g.,}}

\def\etal{{\em et al.}}

\newcommand{\comment}[1]{}

\newcounter{prop}
\newtheorem{proposition}[prop]{Proposition}

\usetikzlibrary{arrows}

\tikzset{
  treenode/.style = {align=center, inner sep=0pt, text centered,
    font=\sffamily},
  arn_n/.style = {treenode, circle, white, font=\sffamily\bfseries, draw=black,
    fill=black, text width=1.4em,solid},
  arn_r/.style = {treenode, circle, black, draw=black, fill=gray,
    text width=1.4em, very thick,solid},
  arn_r2/.style = {treenode, circle, red, draw=gray, 
    text width=1.4em, very thick,solid},
  arn_x/.style = {treenode, rectangle, draw=black,
    minimum width=0.5em, minimum height=0.5em,solid},
    emph/.style={ solid, very thick },
    norm/.style={ dashed, thin}
}

%% file: abs.tex
Decision trees and randomized forests are widely used in computer vision and machine learning. Standard algorithms for decision tree induction optimize the split functions one node at a time according to some splitting criteria. This greedy procedure often leads to suboptimal trees. In this paper, we present an algorithm for optimizing the split functions at all levels of the tree jointly with the leaf parameters, based on a global objective. We show that the problem of finding optimal linear-combination (oblique) splits for decision trees is related to structured prediction with latent variables, and we formulate a convex-concave upper bound on the tree's empirical loss. The run-time of computing the gradient of the proposed surrogate objective with respect to each training exemplar is quadratic in the the tree depth, and thus training deep trees is feasible. The use of stochastic gradient descent for optimization enables effective training with large datasets. Experiments on several classification benchmarks demonstrate that the resulting non-greedy decision trees outperform greedy decision tree baselines.

%% file: intro.tex
\section{Introduction}

Decision trees and forests~\cite{breiman1984classification,
  quinlan1986induction, Breiman01} have a long and rich history in
machine learning~\cite{trevor2009elements, AntonioJamieBook}.  Recent
years have seen an increase in their popularity, owing to their
computational efficiency and applicability to large-scale
classification and regression tasks. A case in point is Microsoft
Kinect where decision trees are trained on millions of exemplars to
enable real-time human pose estimation from depth
images~\cite{ShottonPami13}.

Conventional algorithms for decision tree induction are greedy. They
grow a tree one node at a time following procedures laid out decades
ago by frameworks such as ID3~\cite{quinlan1986induction} and
CART~\cite{breiman1984classification}. While recent work has proposed
new objective functions to guide greedy
algorithms~\cite{NowozinICML12,JancsaryECCV12}, it continues to be the
case that decision tree applications (\eg~\cite{gall2011,
  konukoglu2012}) utilize the same dated methods of tree induction.
Greedy decision tree induction builds a binary tree via a recursive
procedure as follows: beginning with a single node, indexed by $i$, a
split function $s_i$ is optimized based on a corresponding subset of
the training data $\Dset_i$ such that $\Dset_i$ is split into two
subsets, which in turn define the training data for the two children
of the node $i$. The intrinsic limitation of this procedure is that
the optimization of $s_i$ is solely conditioned on $\Dset_i$,
\ie~there is no ability to fine-tune the split function $s_i$ based on
the results of training at lower levels of the tree.  This paper
addresses this limitation by proposing a general framework for
non-greedy learning of the split parameters for tree-based methods.
We focus on binary trees, while extension to $n$-ary trees is
possible.  We show that our joint optimization of the split functions
at different levels of the tree under a global objective not only
promotes cooperation between the split nodes to create more compact
trees, but also leads to better generalization performance.


One of the key contributions of this work is establishing a link
between the decision tree optimization problem and the problem of
structured prediction with latent variables~\cite{YuJ09}.  We present
a novel formulation of the decision tree learning that associates a
binary latent decision variable with each split node in the tree and
uses such latent variables to formulate the tree's empirical loss.
Inspired by advances in structured prediction~\cite{TaskarGK03,
  TsochantaridisHJA04, YuJ09}, we propose a convex-concave upper bound
on the empirical loss. This bound acts as a surrogate objective that
is optimized using stochastic gradient descent (SGD) to find a locally
optimal configuration of the split functions.  One complication
introduced by this particular formulation is that the number of latent
decision variables grows exponentially with the tree depth $d$. As a
consequence, each gradient update will have a complexity of $O(2^d\p)$
for $\p$-dimensional inputs. One of our technical contributions is
showing how this complexity can be reduced to $O(d^2\p)$ by modifying
the surrogate objective, thereby enabling efficient training of deep
trees.

%% file: previous.tex
\vspace*{-.1cm}
\section{Related work}
\vspace*{-.1cm}

Finding optimal split functions at different levels of a decision tree
according to some global objective, such as a regularized empirical
risk, is NP-complete~\cite{HyafilR76} due to the discrete and
sequential nature of the decisions in a tree.  Thus, finding an
efficient alternative to the greedy approach has remained a difficult
objective despite many prior attempts.

Bennett~\cite{bennett1994global} proposes a non-greedy multi-linear
programming based approach for global tree optimization and shows that
the method produces trees that have higher classification accuracy
than standard greedy trees.  However, their method is limited to
binary classification with $0$-$1$ loss and has a high computation
complexity, making it only applicable to trees with few nodes.

The work in \cite{lakshminarayanan2014-mondrian} proposes a means for
training decision forests in an online setting by incrementally
extending the trees as new data points are added.  As opposed to a
naive incremental growing of the trees, this work models the decision
trees with Mondrian Processes.


The Hierarchical Mixture of Experts model~\cite{JordanJacobs94} uses
soft splits rather than hard binary decisions to capture situations
where the transition from low to high response is gradual.  The use of
soft splits at internal nodes of the tree yields a probabilistic model
in which the log-likelihood is a smooth function of the unknown
parameters.  Hence, training based on log-likelihood is amenable to
numerical optimization via methods such as expectation maximization
(EM).  That said, the soft splits necessitate the evaluation of all or
most of the experts for each data point, so much of the computational
advantage of the decision trees are lost.


Murthy and Salzburg~\cite{murthy1995growing} argue that non-greedy
tree learning methods that work by looking ahead are unnecessary and
sometimes harmful. This is understandable since their methods work by
minimizing empirical loss without any regularization, which is prone
to overfitting. To avoid this problem, it is a common practice (see
Breiman~\cite{Breiman01} or Criminisi and
Shotton~\cite{AntonioJamieBook} for an overview) to limit the tree
depth and introduce limits on the number of training instances below
which a tree branch is not extended, or to force a diverse ensemble of
trees (\ie~a decision forest) through the use of
bagging~\cite{Breiman01} or
boosting~\cite{friedman2001greedy}. Bennett and
Blue~\cite{Bennett97asupport} describe a different way to overcome
overfitting by using max-margin framework and the Support Vector
Machines (SVM) at the split nodes of the tree.  Subsequently, Bennett
\etal~\cite{bennett2000enlarging} show how enlarging the margin of
decision tree classifiers results in better generalization
performance.

Our formulation for decision tree induction improves on prior art in a
number of ways. Not only does our latent variable formulation of
decision trees enable efficient learning, but it also handles any general
loss function while not sacrificing the $O(d\p)$ complexity of
inference imparted by the tree structure. Further, our surrogate
objective provides a natural way to regularize the joint optimization
of tree parameters to discourage overfitting.

%% file: formulation.tex
\vspace*{-.1cm}
\section{Problem formulation}
\vspace*{-.1cm}

For ease of exposition, this paper focuses on binary classification
trees, with $\m$ internal (split) nodes, and $\m+1$ leaf (terminal)
nodes. Note that in a binary tree the number of leaves is always one
more than the number of internal (non-leaf) nodes. An input, $\x \in
\Real^\p$, is directed from the root of the tree down through internal
nodes to a leaf node. Each leaf node specifies a distribution over $k$
class labels. Each internal node, indexed by $i \in \{1, \ldots, m\}$,
performs a binary test by evaluating a node-specific split function
$\test_i(\x): \Real^p \to \{-1, +1\}$.  If $\test_i(\x)$ evaluates to
$-1$, then $\x$ is directed to the left child of node $i$. Otherwise,
$\x$ is directed to the right child.  And so on down the tree.  Each
split function $\test_i(\cdot)$, parameterized by a weight vector
$\w_i$, is assumed to be a linear threshold function, \ie~$\test_i(\x)
= \sign(\trans{\w_i} \x)$. We incorporate an offset parameter to
obtain split functions of the form $\sign(\trans{\w_i} \x - b_i)$ by
appending a constant ``$-1$'' to the input feature vector.

Each leaf node, indexed by $j \in \{1, \ldots, m+1\}$, specifies a
conditional probability distribution over class labels, $l \in
\{1,\ldots,k\}$, denoted $p(\y = l \mid j)$.
Leaf distributions
are parametrized with a vector of unnormalized predictive
log-probabilities, denoted $\leaf_j \in \Real^k$, and a 
softmax function; \ie
\begin{equation}
p(\y = l\mid j) ~=~ \frac{\exp \left\{ \leaf_{j[l]}
  \right\}}{\sum_{\alpha=1}^k\exp \left\{ \leaf_{j[\alpha]}
  \right\}}~,
\label{eq:tree-pred-softmax}
\end{equation}
where $\leaf_{j[\alpha]}$ denotes the $\alpha^\mathrm{th}$ element of
vector $\leaf_{j}$.

The parameters of the tree comprise the $m$ internal weight vectors,
$\{\w_i\}_{i=1}^m$, and the $m+1$ vectors of unnormalized
log-probabilities, one for each leaf node, $\{\leaf_j\}_{j=1}^{m+1}$. We
pack these parameters into two matrices $\W \in \Real^{\m \times \p}$
and $\Leaves \in \Real^{(\m+1) \times k}$ whose rows comprise weight
vectors and leaf parameters, \ie~$W \equiv \trans{[\w_1, \ldots,
    \w_\m]}$ and $\Leaves \equiv \trans{[\leaf_{1}, \ldots,
    \leaf_{m+1} ]}$. Given a dataset of input-output pairs, $\Dset
\equiv \{ \x_z, \y_z \}_{z=1}^n$, where $\y_z \in \{1, \ldots, k \}$
is the ground truth class label associated with input $\x_z \in
\Real^\p$, we wish to find a joint configuration of oblique splits
$\W$ and leaf parameters $\Leaves$ that minimize some measure of
misclassification loss on the training dataset.  Joint optimization of the
split functions and leaf parameters according to a global objective is
known to be extremely challenging~\cite{HyafilR76} due to the discrete
and sequential nature of the splitting decisions within the tree.

One can evaluate all of the split functions, for every internal node
of the tree, on an input $\x$ by computing $\sign(W \x)$, where
$\sign(\cdot)$ is the element-wise sign function. One key idea that
helps linking decision tree learning to latent structured prediction
is to think of an $\m$-bit vector of potential split decisions,
\eg~$\h = \sign(W \x) \in \{-1, +1\}^\m$, as a latent variable. Such a
latent variable determines the leaf to which a data point is directed,
and then classified using the leaf parameters. To formulate the loss
for an input-output pair, $(\x,\y)$, we introduce a tree navigation
function $\f{\m} :
\Bin{\m} \to \Ind{\m + 1}$ that maps an $\m$-bit sequence of split
decisions ($\Bin{\m} \equiv \{-1, +1\}^\m$) to an indicator vector
that specifies a $1$-of-$(\m+1)$ encoding. Such an indicator vector is
only non-zero at the index of the selected leaf. \figref{fig:fexample}
illustrates the tree navigation function for a tree with $3$ internal
nodes.

\begin{figure}
\input{figs/treenav.tex}
\caption{The binary split decisions in a decision tree with $\m = 3$
  internal nodes can be thought as a binary vector $\h = \trans{[h_1,
      h_2, h_3]}$. Tree navigation to reach a leaf can be expressed in
  terms of a function $\f{\m}(\h)$. The selected leaf parameters can be
  expressed by $\leaf = \trans{\Leaves}\f{\m}(\h)$.}
\label{fig:fexample}
\vspace*{-.2cm}
\end{figure}

Using the notation developed above, $\leaf = \trans{\Leaves}
\f{\m}(\sign(W \x))$ represents the parameters corresponding to the
leaf to which $\x$ is directed by the split functions in $W$. A
generic loss function of the form $\ell(\leaf, \y)$ measures the
discrepancy between the model prediction based on $\leaf$ and an
output~$\y$. For the softmax model given by
\eqref{eq:tree-pred-softmax}, a natural loss is the negative log
probability of the correct label, referred to as {\em log loss},
\begin{equation}
\ell(\leaf, \y) ~=~ \ell_{\log}(\leaf, \y)~ =~ -\leaf_{[\y]} +
\log\bigg(\sum_{\beta = 1}^{k} \exp(\leaf_{[\beta]})\bigg)~.
\end{equation}
For regression tasks, when $\vec{\y} \in \Real^q$, and the value of
$\leaf \in \Real^q$ is directly emitted as the model prediction, a natural
choice of $\ell$ is squared loss,
\begin{equation}
\ell(\leaf, \vec{\y}) ~=~ \ell_{\mathrm{sqr}}(\leaf, \vec{\y}) ~=~
\|\leaf - \vec{\y}\|^2~.
\end{equation}
One can adopt other forms of loss within our decision tree learning
framework as well. The goal of learning is to find $\W$ and $\Leaves$
that minimize empirical loss, for a given training set $\Dset$, that
is,
\begin{equation}
\calL(\W,\Leaves;\Dset)~ =~ \sumxy \ell \left( \trans{\Leaves}
\f{\m}(\sign(\W\x)) , y \right)~.
\label{eq:emp-risk-dtree}
\end{equation}
Direct global optimization of empirical loss $\calL(\W,\Leaves;\Dset)$
with respect to $\W$ is challenging.  It is a discontinuous and
piecewise-constant function of $\W$.  Furthermore, given an input
$\x$, the navigation function $\f{\m}(\cdot)$ yields a leaf parameter
vector based on a sequence of binary tests, where the results of the
initial tests determine which subsequent tests are performed.  It is
not clear how this dependence of binary tests should be formulated.

%% file: figs/treenav.tex
\centering
\begin{minipage}{0.45\textwidth}
\centering
\begin{tikzpicture}[->,>=stealth',level/.style={sibling distance = 2.5cm/#1,
  level distance = 1.5cm}] 
\begin{scope}[scale = .6]
\node [arn_n,label={$h_1$}] {+1}
    child{ node [arn_n,label={$h_2$}] {-1}  
           child{ node [arn_r2, label=below:$\leaf_1$] {} edge from parent[emph]}
	child{ node [arn_r2, label=below:$\leaf_2$] {} edge from parent[norm]}
           edge from parent[norm]	
    }
    child{ node [arn_n,label={$h_3$}] {+1}
	child{ node [arn_r2, label=below:$\leaf_3$] {} edge from parent[norm]}
	child{ node [arn_r, label=below:$\leaf_4$] {} edge from parent[emph]}
           edge from parent[emph]
    }
; 
\end{scope}
\end{tikzpicture}
\begin{equation*}
\f{3}(\trans{[+1, -1, +1]}) \:=\: \trans{[0, 0, 0, 1]} \:=\: \1{}_{4}
\end{equation*}
\begin{equation*}
\leaf \:=\: \trans{\Leaves}\f{3}(\h) = \leaf_{4}
\end{equation*}
\end{minipage}
\hspace*{.7cm}
\begin{minipage}{0.45\textwidth}
\centering
\begin{tikzpicture}[->,>=stealth',level/.style={sibling distance = 2.5cm/#1,
  level distance = 1.5cm}]
\begin{scope}[scale = .6]
\node [arn_n,label={$h_1$}] {-1}
    child{ node [arn_n,label={$h_2$}] {+1}
           child{ node [arn_r2, label=below:$\leaf_1$] {} edge from parent[norm]}
	child{ node [arn_r, label=below:$\leaf_2$] {} edge from parent[emph]}
           edge from parent[emph]	
    }
    child{ node [arn_n,label={$h_3$}] {+1}
	child{ node [arn_r2, label=below:$\leaf_3$] {} edge from parent[norm]}
	child{ node [arn_r2, label=below:$\leaf_4$] {} edge from parent[emph]}
           edge from parent[norm]
    }
; 
\end{scope}
\end{tikzpicture}
\begin{equation*}
\f{3}(\trans{[-1, +1, +1]}) \:=\: \trans{[0, 1, 0, 0]} \:=\: \1{}_{2}
\end{equation*}
\begin{equation*}
\leaf \:=\: \trans{\Leaves}\f{3}(\h) \:=\: \leaf_{2}
\end{equation*}
\end{minipage}

%% file: learning.tex
\vspace*{-.1cm}
\section{Decision trees and structured prediction}
\vspace*{-.1cm}

To overcome the intractability in the optimization of~$\calL$, we
develop a piecewise smooth upper bound on empirical loss.  Our upper
bound is inspired by the formulation of structured prediction with
latent variables~\cite{YuJ09}.  A key
observation that links decision tree learning to structured
prediction, is that one can re-express $\sign(\W\x)$ in terms of a
latent variable $\h$. That is,
\begin{equation}
\sign(\W\x) ~=~ \argmax{\h \in \Bin{\m}}(\trans{\h}\W\x)~.
\label{eq:sign-as-max}
\end{equation}
In this form, decision tree's split functions implicitly map an input
$\x$ to a binary vector $\h$ by maximizing a score function
$\trans{\h} \W \x$, the inner product of $\h$ and $\W\x$. One can
re-express the score function in terms of a more familiar form of a
joint feature space on $\h$ and $\x$, as $\trans{\vec{w}}\phi(\h,
\x)$, where $\phi(\h, \x) = \operatorname{vec}{(\h\trans{\x})}$, and
$\w = \operatorname{vec}{(\W)}$.  Previously,
Norouzi~\etal~\cite{NorouziFICML11,NorouziFSNIPS12} used the same
re-formulation (\ref{eq:sign-as-max}) of linear threshold functions to
learn binary similarity preserving hash functions.

Given \eqref{eq:sign-as-max}, we re-express empirical loss as,
\begin{equation}
\begin{aligned}
\calL (\W, \Leaves ; \Dset) = \sumxy \ell(\trans{\Leaves}\f{m}(\widehat{\h}(\x)), \y)&~,\\
\label{eq:sp-emp-loss}
\text{\em where} \hspace*{.4cm} \widehat{\h}(\x) = ~\argmax{\h \in \Bin{\m}} (\trans{\h} \W \x )~.&
\end{aligned}
\end{equation}
This objective resembles the objective functions used in structured
prediction, and since we do not have {\em a priori} access to the
ground truth split decisions, $\widehat{\h}(\x)$, this problem is a
form of structured prediction with latent variables.

\vspace*{-.1cm}
\section{Upper bound on empirical loss}
\vspace*{-.1cm}

We develop an upper bound on loss for an input-output pair, $(\x, \y)$,
which takes the form,
\begin{equation}
\ell(\trans{\Leaves} \f{m}(\sign(\W \x)), \y) ~\, \le ~\, \max_{\g \in
  \Bin{\m}}\Big(\trans{\g}\W\x \, + \, \ell(\trans{\Leaves} \f{m}(\g),
\y)\Big) ~-~ \max_{\h \in \Bin{\m}}(\trans{\h}\W\x) ~ .
\label{eq:upper-tree-loss}
\end{equation}
To validate the bound, first note that the second term on the RHS is
maximized by $\h = \widehat{\h}(\x) = \sign(\W \x)$.  Second, when $\g
= \widehat{\h}(\x)$, it is clear that the LHS equals the RHS.
For all other values of $\g$, the RHS can only get larger
than when $\g = \widehat{\h}(\x)$ because of the max operator. Hence,
the inequality holds. An algebraic proof of \eqref{eq:upper-tree-loss}
is presented in the Appendix.

In the context of structured prediction, the first term of the upper
bound, \ie~the maximization over $\g$, is called {\em loss-augmented
  inference}, as it augments the standard inference problem, \ie~the
maximization over $\h$, with a loss term.  Fortunately, the
loss-augmented inference for our decision tree learning formulation
can be solved exactly, as discussed below.

It is also notable that the loss term on the LHS of
\eqref{eq:upper-tree-loss} is invariant to the scale of $\W$, but the
upper bound on the right side of \eqref{eq:upper-tree-loss} is not. As
a consequence, as with binary SVM and margin-rescaling formulations
of structural SVM~\cite{TsochantaridisHJA04}, we introduce a
regularizer on the norm of $\W$ when optimizing the bound. To justify
the regularizer, we discuss the effect of the scale of $\W$ on the bound.

\comment {{

the implicit margin constraints in \eqref{eq:upper-tree-loss} are
easier to satisfy when the scale of $\W$ is large.

}}

\begin{proposition}
\label{prop:tightness}
The upper bound on the loss becomes tighter as a constant multiple of
$\W$ increases, i.e.,~for $a > b > 0$:
\begin{equation}
\begin{aligned}
\max_{\g \in \Bin{\m}}\Big(a\trans{\g}\W\x + \ell(\trans{\Leaves}
\f{m}(\g), \y)\Big) - \max_{\h \in \Bin{\m}} ( a\trans{\h}\W\x ) &~~\le
\\ \max_{\g \in \Bin{\m}}\Big( b\trans{\g}\W\x +
\loss(\trans{\Leaves} \f{m}(\g), \y)\Big) &- \max_{\h \in \Bin{\m}} (
b\trans{\h}\W\x ).
\end{aligned}
\end{equation}
\begin{proof}
Please refer to the Appendix for the proof.
\end{proof}
\end{proposition}


In the limit, as the scale of $\W$ approach $+\infty$, the loss term
$\loss(\trans{\Leaves} \f{m}(\g), \y)$ becomes negligible compared to
the score term $\trans{\g}\W\x$.  Thus, the solutions to
loss-augmented inference and inference become almost identical, except
when an element of $\W\x$ is very close to $0$. Thus, even though a
larger $\lVert\W\rVert$ yields a tighter bound, it makes the bound
approach the loss itself, and therefore becomes nearly
piecewise-constant, which is hard to optimize. In fact, based on
\propref{prop:tightness}, one easy way to decrease the upper bound is
to increase the norm of $\W$, which does not affect the loss.

Our experiments indicate that when the norm of $\W$ is regularized, a
lower value of the loss at both training and validation time can be
achieved. We therefore constrain the norm of $\W$ to obtain an
objective with better behavior and generalization. Since each row of
$\W$ acts independently in a decision tree in the split functions, it
is reasonable to constrain the norm of each row independently. Summing
over the bounds for different training pairs and constraining the norm
of rows of $\W$, we obtain the following optimization problem, called
the {\em surrogate} objective:
\begin{equation}
\begin{aligned}
\textrm{minimize}~\calL' (\W, \Leaves ; \Dset)~ = \sumxy \bigg(
\max_{\g \in \Bin{\m}}\Big( & \trans{\g}\W\x + \ell(\trans{\Leaves}
\f{m}(\g), \y) \Big) - \max_{\h \in \Bin{\m}}( \trans{\h}\W\x) \bigg)
\\ \text{\em s.t.} \quad \| \w_i \|^2 &\le \nu ~~~~ \text{\em for
  all~~} i \in \{1, \ldots ,\m \}~,
\end{aligned}
\label{eq:upper-dtree-emploss}
\end{equation}
where $\nu \in \Real^+$ is a regularization parameter and $\w_{i}$ is
the $\th{i}$ row of $\W$.  For all values of $\nu$, we have $\calL(\W,
\Leaves ; \Dset) \le \calL' (\W, \Leaves ; \Dset)$. Instead of using
the typical Lagrange form for regularization, we employ hard
constraints to enable sparse gradient updates of the rows of $\W$,
since as explained below, the gradients for most rows of $\W$ are zero
at each step of training.


%% file: opt.tex
\vspace*{-.1cm}
\section{Optimizing the surrogate objective}
\vspace*{-.1cm}
\label{sec:optim}

Even though minimizing the surrogate objective of
\eqref{eq:upper-dtree-emploss} entails non-convex optimization,
$\calL'(\W, \Leaves; \Dset)$ is much better behaved than empirical
loss in \eqref{eq:emp-risk-dtree}. $\calL'(\W, \Leaves; \Dset)$ is
piecewise linear and convex-concave in $\W$, and the constraints on
$\W$ define a convex set.


{\bf Loss-augmented inference.} To evaluate and use the
surrogate objective in \eqref{eq:upper-dtree-emploss} for
optimization, we must solve a {\em loss-augmented inference} problem
to find the binary code that maximizes the sum of the score and loss
terms:
\begin{equation}
\widehat{\g}(\x) ~= ~ \argmax{\g \in \Bin{\m}} \, \left( \,
\trans{\g}\W\x + \loss(\trans{\Leaves} \f{\m}(\g), \, \y) \right).
\label{eq:loss-aug-inf-dtree}
\end{equation}
An observation that makes this optimization tractable is that
$\f{\m}(\g)$ can only take on $\m\!+\!1$ distinct values, which
correspond to terminating at one of the $\m\!+\!1$ leaves of the tree
and selecting a leaf parameter from $\{\leaf_j\}_{j=1}^{m+1}$.
Fortunately, for any leaf index $j \in \{1, \ldots, m\!+\!1\}$, we can
solve
\begin{equation}
\argmax{\g \in \Bin{\m}} \, \left( \, \trans{\g}\W\x + \loss(\leaf_j,
\y) \right)~~~~~\text{\em s.\,t.}~~~~~\f{\m}(\g) = \1{}_j~,
\label{eq:loss-aug-inf2-dtree}
\end{equation}
efficiently. Note that if $\f{\m}(\g) = \1{}_j$, then $\trans{\Leaves}
\f{\m}(\g)$ equals the $\th{j}$ row of $\Leaves$, \ie~$\leaf_j$. To
solve \eqref{eq:loss-aug-inf2-dtree} we need to set all of the binary
bits in $\g$ corresponding to the path from the root to the leaf $j$
to be consistent with the path direction toward the leaf $j$.
However, bits of $\g$ that do not appear on this path have no effect
on the output of $\f{\m}(\g)$, and all such bits should be set based
on $\g_{[i]} = \sign(\trans{\w_i}\x)$ to obtain maximum
$\trans{\g}\W\x$. Accordingly, we can essentially ignore the
off-the-path bits by subtracting $\trans{\sign(\W\x)}W\x$ from
\eqref{eq:loss-aug-inf2-dtree} to obtain,
\begin{equation}
\argmax{\g \in \Bin{\m}} \, \left( \trans{\g}\W\x + \loss(\leaf_j, \y) \right) ~=~ 
\argmax{\g \in \Bin{\m}} \, \left( \trans{\big(\g - \sign(W\x)\big)}\W\x + \loss(\leaf_j, \y) \right)~.
\label{eq:loss-aug-inf3-dtree}
\end{equation}
Note that $\trans{\sign(\W\x)}W\x$ is constant in $\g$, and this
subtraction zeros out all bits in $\g$ that are not on the path to the
leaf $j$. So, to solve \eqref{eq:loss-aug-inf3-dtree}, we only need to
consider the bits on the path to the leaf $j$ for which
$\sign(\trans{\w_i}\x)$ is not consistent with the path direction.
Using a single depth-first search on the decision tree, we can solve
\eqref{eq:loss-aug-inf2-dtree} for every $j$, and among those, we pick
the one that maximizes \eqref{eq:loss-aug-inf2-dtree}.


The algorithm described above is $O(\m\p) \subseteq O(2^\depth\p)$,
where $\depth$ is the tree depth, and we require a multiple of $\p$
for computing the inner product $\w_i\x$ at each internal node
$i$. This algorithm is not efficient for deep trees, especially as we
need to perform loss-augmented inference once for every stochastic
gradient computation.  In what follows, we develop an alternative more
efficient formulation and algorithm with time complexity of
$O(\depth^2\p)$.

{\bf Fast loss-augmented inference.} To develop a faster
loss-augmented inference algorithm, we formulate a slightly different
upper bound on the loss, \ie~
\begin{equation}
\loss(\trans{\Leaves} \f{\m}(\sign(\W \x)), \y) ~ \le~ 
\max_{\g \in \Bset_1(\sign(\W\x))} \!\!\big( \trans{\g}\W\x + \loss(\trans{\Leaves} \f{\m}(\g), \y) \big) 
- \max_{\h \in \Bin{\m}} \big( \trans{\h}\W\x \big)~,
\label{eq:upper-tree-loss2}
\end{equation}
where $\Bset_1(\sign(\W\x))$ denotes the Hamming ball of radius $1$
around $\sign(\W\x)$, \ie~$\Bset_1(\sign(\W\x)) \equiv \{\g \in
\Bin{\m} ~|~ \lVert \g - \sign(\W\x) \rVert_H \le 1 \}$, hence $\g \in
\Bset_1(\sign(\W\x))$ implies that $\g$ and $\sign(\W\x)$ differ in at
most one bit. The proof of \eqref{eq:upper-tree-loss2} is identical to
the proof of \eqref{eq:upper-tree-loss}. The key benefit of this new
formulation is that loss-augmented inference with the new bound is
computationally efficient.  Since $\widehat{\g}$ and $\sign(\W\x)$
differ in at most one bit, then $\f{\m}(\widehat{\g})$ can only take
$d+1$ distinct values.  Thus we need to evaluate
\eqref{eq:loss-aug-inf3-dtree} for at most $d+1$ values of $j$,
requiring a running time of $O(d^2p)$.


{\bf Stochastic gradient descent (SGD).} A reasonable approach to
minimizing \eqref{eq:upper-dtree-emploss} uses stochastic gradient
descent (SGD), the steps of which are outlined in Alg~\ref{alg:sgd}.
Here, $\eta$ denotes the learning rate, and $\tau$ is the number of
optimization steps.  Line $6$ corresponds to a gradient update in
$\W$, which is supported by the fact that $\frac{\partial}{\partial
  \W} {\trans{\h} \W \x} = \h\trans{\x}$.  Line $8$ performs
projection back to the feasible region of $\W$, and Line $10$ updates
$\Leaves$ based on the gradient of the loss. Our implementation
modifies Alg~\ref{alg:sgd} by adopting common SGD tricks, including
the use of momentum and mini-batches.


\input{algorithm-sgd.tex}

{\bf Stable SGD (SSGD).}  Even though Alg~\ref{alg:sgd} achieves good
training and test accuracy relatively quickly, we observe that after
several gradient updates some of the leaves may end up not being
assigned to any data points and hence the full tree capacity may not
be exploited. We call such leaves {\em inactive} as opposed to {\em
  active} leaves that are assigned to at least one training data
point. An inactive leaf may become active again, but this rarely
happens given the form of gradient updates. To discourage abrupt
changes in the number of inactive leaves, we introduce a variant of
SGD, in which the assignments of data points to leaves are fixed for a
number of gradient update steps. Thus, the bound is optimized with
respect to a set of data point to leaf assignment constraints.  When
the improvement in the bound becomes negligible the leaf assignment
variables are updated, followed by another round of optimization of
the bound. We call this algorithm {\em Stable SGD (SSGD)} because it
changes the assignment of data points to leaves more conservatively
than SGD. Let $a(\x)$ denote the $1$-of-$(\m+1)$ encoding of the leaf
to which a data point $\x$ should be assigned to. Then, SSGD with fast
loss-augmented inference relies on the following upper bound on loss,
\begin{equation}
\loss(\trans{\Leaves} \f{\m}(\sign(\W \x)), \y) ~ \le~ 
\max_{\g \in \Bset_1(\sign(\W\x))} \!\!\big( \trans{\g}\W\x + \loss(\trans{\Leaves} \f{\m}(\g), \y) \big) 
- \max_{\h \in \Bin{\m} \mid f(\h) = a(\x)} \big( \trans{\h}\W\x \big)~.
\label{eq:upper-tree-loss3}
\end{equation}
One can easily verify that the RHS of \eqref{eq:upper-tree-loss3} is
larger than the RHS of \eqref{eq:upper-tree-loss2}, hence the inequality.

{\bf Computational complexity.}  To analyze the computational
complexity of each SGD and SSGD step, we note that Hamming distance
between $\widehat{\g}$ (defined in \eqref{eq:loss-aug-inf-dtree}) and
$\widehat{\h} = \sign(\W\x)$ is bounded above by the depth of the tree
$d$. This is because only those elements of $\widehat{\g}$
corresponding to the path to a selected leaf can differ from
$\sign(\W\x)$. Thus, for SGD the expression $ (\widehat{\g} -
\widehat{\h}) \,\trans{\x}$ needed for Line 6 of Alg~\ref{alg:sgd} can
be computed in $O(dp)$, if we know which bits of $\widehat{\h}$ and
$\widehat{\g}$ differ.  Accordingly, Lines 6 and 7 can be performed in
$O(dp)$.  The computational bottleneck is the loss augmented inference
in Line 5.  When fast loss-augmented inference is performed in
$O(d^2p)$ time, the total time complexity of gradient update for both
SGD and SSGD becomes $O(d^2p + k)$, where $k$ is the number of labels.

%% file: algorithm-sgd.tex
\begin{algorithm}[t]
\caption{
  \label{alg:sgd}
  Stochastic gradient descent (SGD) algorithm for non-greedy
  decision tree learning.
}
\begin{minipage}{0.75\textwidth}
\begin{algorithmic}[1]
\STATE Initialize $\iterz{W}$ and $\iterz{\Leaves}$ using greedy procedure
\FOR{$t=0$ to $\tau$}
\STATE Sample a pair $(\x, \y)$ uniformly at random from $\Dset$
\STATE $\widehat{\h} \leftarrow \sign(\itert{\W}\x)$
\STATE $\widehat{\g} \leftarrow \operatorname{argmax}_{\g \in \Hset} \, \left\{ \trans{\g}\itert{\W}\x + \loss(\trans{\Leaves} f(\g), \y) \right\}$
\STATE $\itertph{\W} \leftarrow \itert{\W} - \eta \ \widehat{\g}\trans{\x} + \eta \ \widehat{\h}\trans{\x}$\\[.1cm]
\FOR{$i=1$ to $m$}
\STATE $\itertpp{\W}_{\ i,\any} \leftarrow \min \, \Bigl\{ 1, {\footnotesize \slfrac{\sqrt{\nu}}{\bigl\lVert \itertph{\W}_{~i,\any} \bigr\rVert_2}}  \Bigr\} \ \itertph{\W}_{\ i,\any}$
\ENDFOR
\STATE $\itertpp{\Leaves} \leftarrow \itert{\Leaves} - \eta \, \frac{\partial}{\partial \Leaves} \loss(\trans{\Leaves} f(\widehat{\g}), \y){\big \vert}_{\Leaves=\itert{\Leaves}}$
\ENDFOR
\vspace{.1cm}
\end{algorithmic}
\end{minipage}
\end{algorithm}

%% file: expts.tex
\section{Experiments}





Experiments are conducted on several benchmark datasets from
LibSVM~\cite{libsvm} for multi-class classification, namely {\em
  SensIT}, {\em Connect4}, {\em Protein}, and {\em MNIST}. We use the
provided train, validation, test sets when available. If such splits
are not provided, we use a random $80\% / 20\%$ split of the training
data for train and validation sets and a random $64\% / 16\% / 20\%$
split for train, validation, test sets.

\input{figs/height/height.tex}

We compare our method for {\em non-greedy} learning of oblique trees
with several greedy baselines, including conventional {\em
  axis-aligned} trees based on information gain, {\em OC1} oblique
trees~\cite{murthy1995growing} that use coordinate descent for
optimization of the splits, and {\em random} oblique trees that select
the best split function from a set of randomly generated hyperplanes
based on information gain. We also compare with the results of {\em
  CO2}~\cite{norouzi2015co2}, which is a special case of our upper
bound approach applied greedily to trees of depth $1$, one node at a
time. Any base algorithm for learning decision trees can be augmented
by post-training pruning \cite{mingers1989-pruning}, or building
ensembles with bagging~\cite{Breiman01} or
boosting~\cite{friedman2001greedy}. However, the key differences
between non-greedy trees and baseline greedy trees become most
apparent when analyzing individual trees.  For a single tree the major
determinant of accuracy is the size of the tree, which we control by
changing the maximum tree depth.

\figref{fig:height} depicts test and training accuracy for non-greedy
trees and four other baselines as function of tree depth. We evaluate
trees of depth $6$ up to $18$ at depth intervals of $2$. The
hyper-parameters for each method are tuned for each depth
independently. While the absolute accuracy of our non-greedy trees
varies between datasets, a few key observations hold for all cases.
First, we observe that non-greedy trees achieve the best test
performance across tree depths across multiple datasets. Second,
trees trained using our non-greedy approach seem to be less
susceptible to overfitting and achieve better generalization
performance at various tree depths. As described below, we think that
the norm regularization provides a principled way to tune the
tightness of the tree's fit to the training data. Finally, the
comparison between non-greedy and CO2~\cite{norouzi2015co2} trees
concentrates on the non-greediness of the algorithm, as it compares
our method with its simpler variant, which is applied greedily one
node at a time. We find that in most cases, the non-greedy
optimization helps by improving upon the results of CO2.


\definecolor{magenta}{rgb}{1,0,1}
\newlength\figh

\begin{figure*}[t]
\input{tikz/valfig.tex}%
\caption{ The effect of $\nu$ on the structure of the trees trained by
  MNIST.  A small value of $\nu$ prunes the tree to use far fewer
  leaves than an axis-aligned baseline used for initialization (dotted
  line).  }
\vspace*{-.3cm}
\label{fig:effect-nu}
\end{figure*}



\input{figs/timing/timing.tex}

A key hyper-parameter of our method is the regularization constant
$\nu$ in \eqref{eq:upper-dtree-emploss}, which controls the tightness
of the upper bound. With a small $\nu$, the norm constraints force the
method to choose a $W$ with a large margin at each internal node. The
choice of $\nu$ is therefore closely related to the generalization of
the learned trees. As shown in \figref{fig:effect-nu}, $\nu$ also
implicitly controls the degree of pruning of the leaves of the tree
during training.
We train multiple trees for different values of $\nu \in
\{0.1,1,4,10,43,100\}$, and we pick the value of $\nu$ that produces
the tree with minimum validation error. We also tune the choice of the
SGD learning rate, $\eta$, in this step.  Such $\nu$ and $\eta$ are
used to build a tree using the union of both the training and
validation sets, which is evaluated on the test set.


To build non-greedy trees, we initially build an axis-aligned tree
with split functions that threshold a single feature, optimized using
conventional procedures that maximize information gain.  The
axis-aligned split is used to initialize a \emph{greedy} variant of
the tree training procedure, called CO2~\cite{norouzi2015co2}.  This
provides initial values for $\W$ and $\Theta$ for the non-greedy
procedure.

\figref{fig:timing} shows an empirical comparison of training time for
SGD with loss-augmented inference and fast loss-augmented
inference. As expected, run-time of SGD with loss-augmented inference
exhibits exponential growth with deep trees whereas its fast variant
is much more scalable. We expect to see better speedup factors for
larger datasets. Connect4 only has $55,000$ training points.




%% file: figs/height/height.tex
\begin{figure}
  \small
  \centering

  \newlength\figw
  \setlength{\figw}{0.23\linewidth}
  
  \newlength\fighheight
  \setlength{\fighheight}{2.95cm}
  
  \begin{tabular}{@{}c@{}c@{}c@{}c@{}}
    ~~~SensIT & Connect4 & Protein & MNIST \\
    \includegraphics[height=\fighheight]{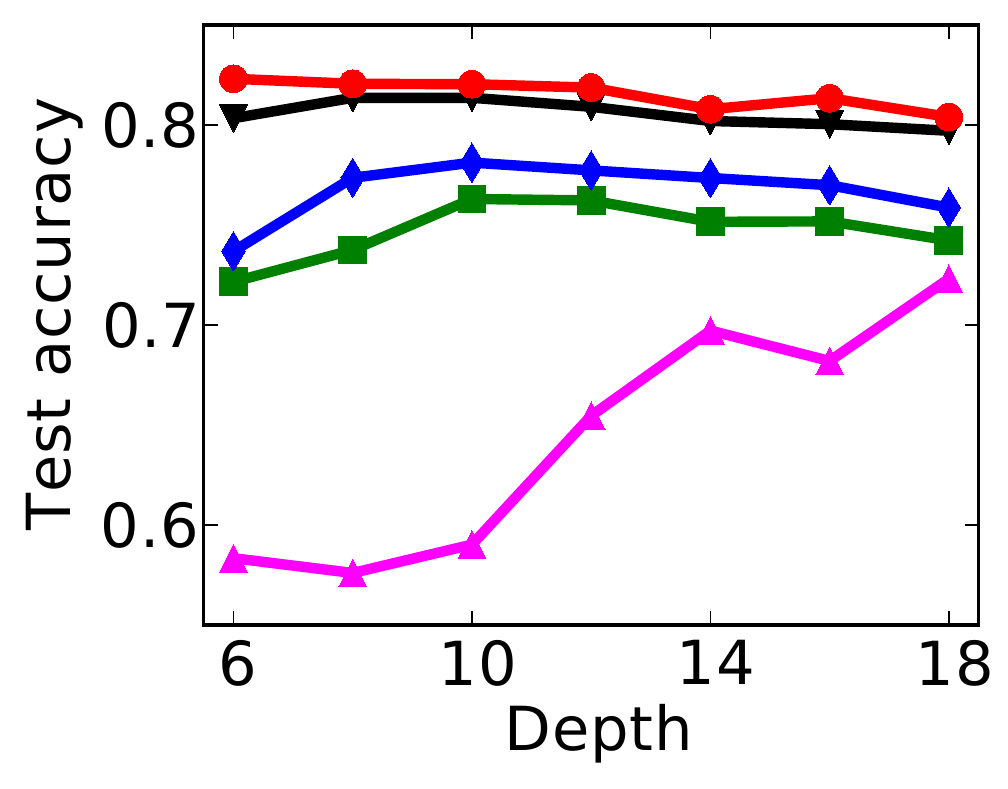}
    & \includegraphics[height=\fighheight]{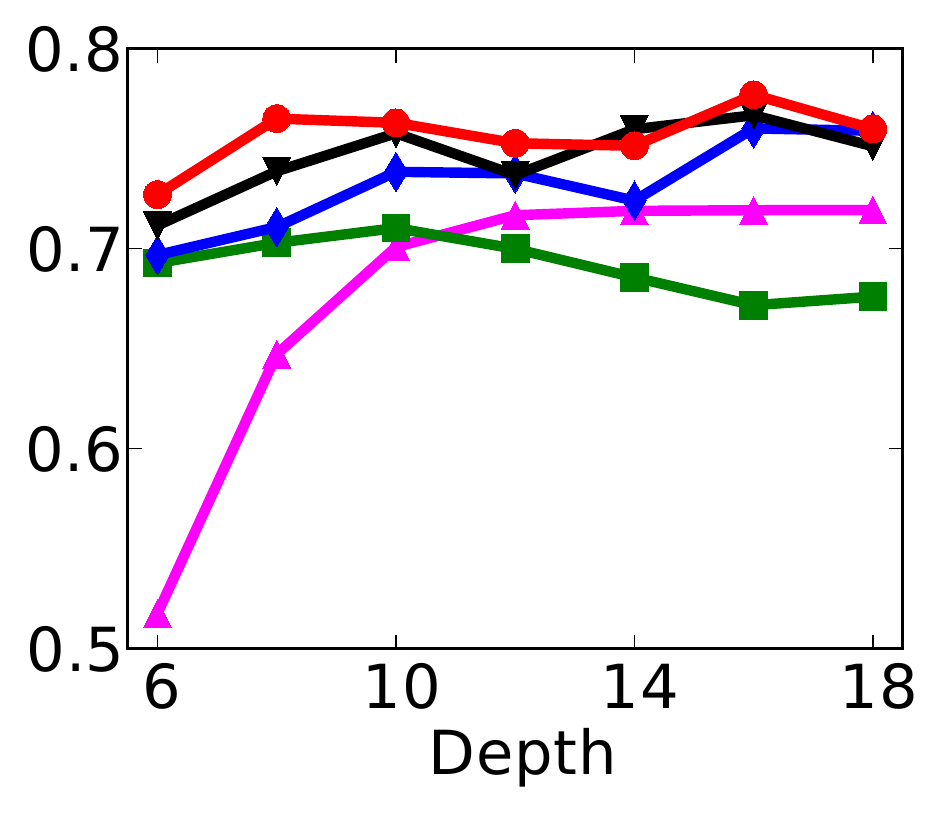}
    & \includegraphics[height=\fighheight]{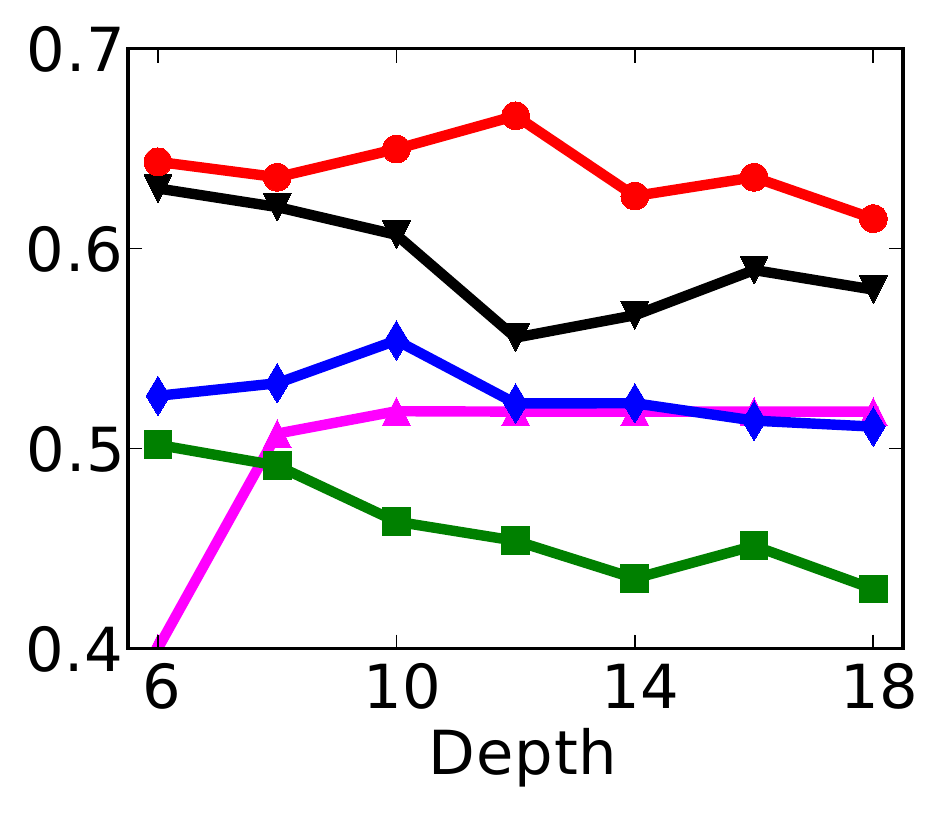}
    & \includegraphics[height=\fighheight]{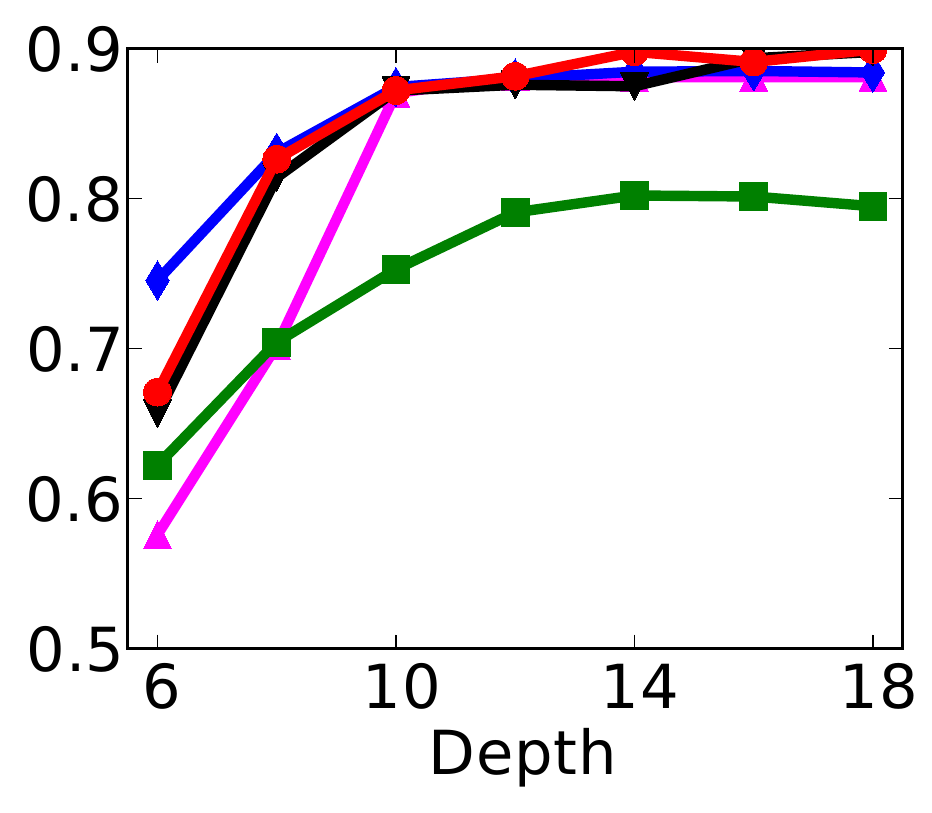} \\
    \includegraphics[height=\fighheight]{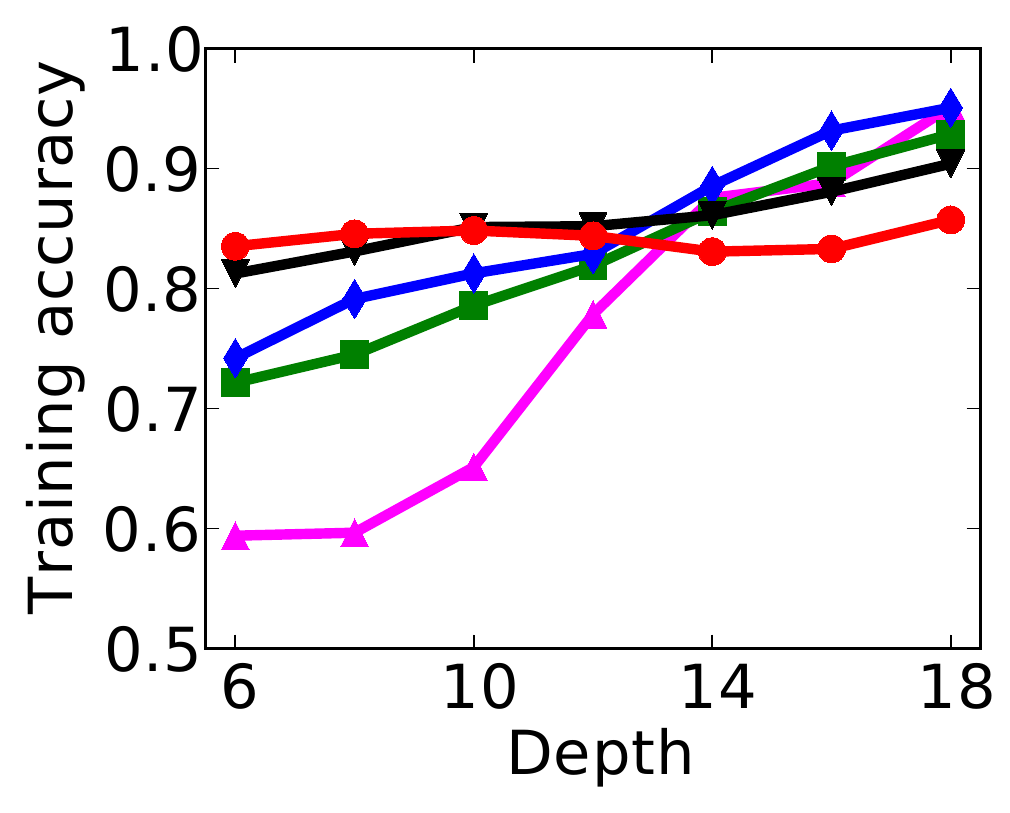}
    & \includegraphics[height=\fighheight]{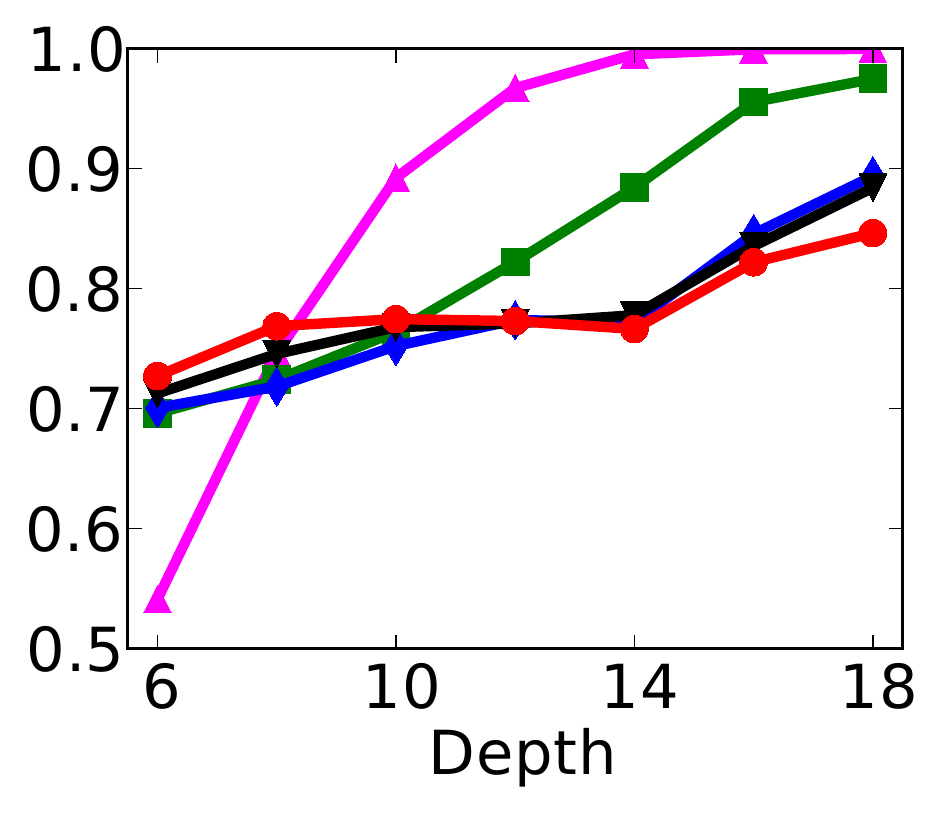}
    & \includegraphics[height=\fighheight]{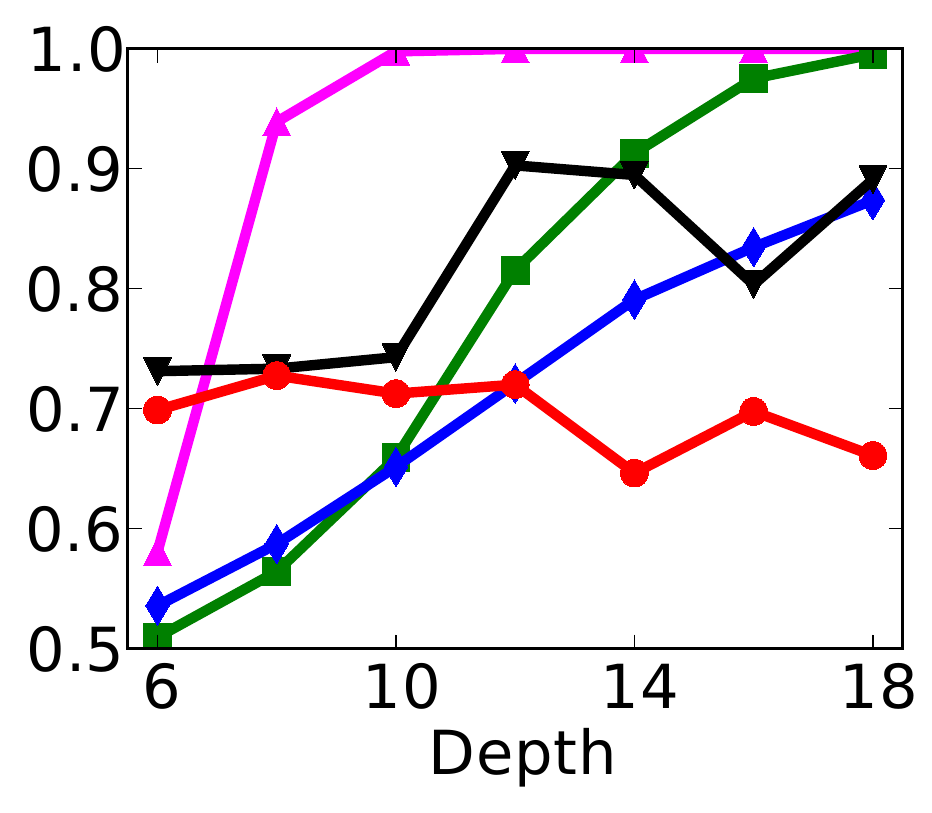}
    & \includegraphics[height=\fighheight]{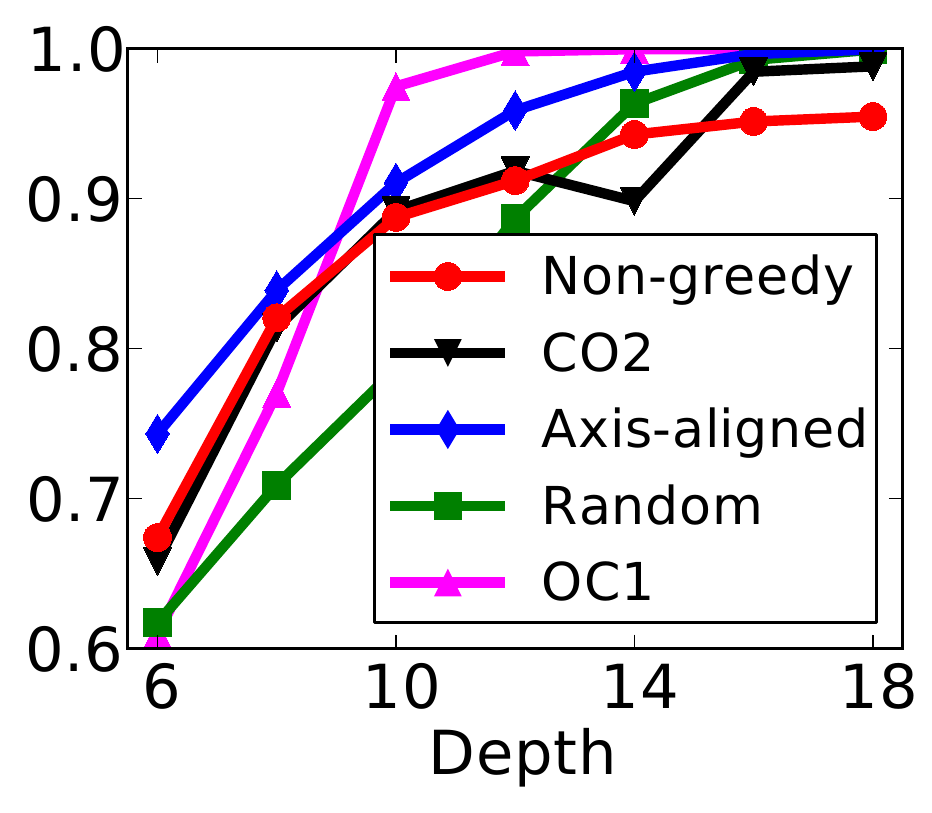}
  \vspace*{-.2cm}
  \end{tabular}
  \caption{ Test and training accuracy of a single tree as a function
    of tree depth for different methods. Non-greedy trees achieve
    better test accuracy throughout different depths. Non-greedy
    exhibit less vulnerability to overfitting.}
  \label{fig:height}
  \vspace*{-.2cm}
\end{figure}

%% file: tikz/valfig.tex
\footnotesize
\setlength\figh{2.8cm}

\begin{minipage}[ht]{0.33\linewidth}
\centering

\begin{tikzpicture}[scale=.9]
\begin{axis}[
xmode=log,
width=1.2\figh,
height=.7\figh,
scale only axis,
xmin=1, xmax=1000,
xlabel={Regularization parameter $\nu$ ($\log$)},
ymin=0, ymax=4000,
ylabel={Num. active leaves},
ymajorgrids,
legend style={at={(0.03,0.03)},anchor=south west,draw=black,fill=white}]
\addplot [
color=green!60!black,
solid,
line width=1.25pt,
mark size=1.4pt,
mark=*,
mark options={solid,fill=green!60!black}
]
coordinates{(    1,   296) ( 3.16,   413) (   10,   477) (31.62,   580) (  100,   635) (316.22,   730) ( 1000,   827)};

\addplot [
color=green!60!black,
dashed,
line width=1.25pt,
mark size=0pt,
mark=*,
mark options={solid,fill=blue}
]
coordinates{(    1, 924) ( 1000, 924)};

\end{axis}
\end{tikzpicture}%

Tree depth $d$ =10
\end{minipage}~\hspace*{.3cm}~\begin{minipage}[ht]{0.30\linewidth}  
\centering

\begin{tikzpicture}[scale=.9]
\begin{axis}[
xmode=log,
width=1.2\figh,
height=.7\figh,
scale only axis,
xmin=1, xmax=1000,
xlabel={Regularization parameter $\nu$ ($\log$)},
ymin=0, ymax=4000,
ymajorgrids,
legend style={at={(0.03,0.03)},anchor=south west,draw=black,fill=white}]
\addplot [
color=green!60!black,
solid,
line width=1.25pt,
mark size=1.4pt,
mark=*,
mark options={solid,fill=green!60!black}
]
coordinates{(    1,   393) ( 3.16,   612) (   10,   900) (31.62,  1166) (  100,  1493) (316.22,  2262) ( 1000,  2485)};

\addplot [
color=green!60!black,
dashed,
line width=1.25pt,
mark size=0pt,
mark=*,
mark options={solid,fill=blue}
]
coordinates{(    1, 2947) ( 1000, 2947)};

\end{axis}
\end{tikzpicture}%

Tree depth $d$ =13
\end{minipage}~\hspace*{.3cm}~\begin{minipage}[ht]{0.30\linewidth}  
\centering
\hspace*{.15cm}
\begin{tikzpicture}[scale=.9]
\begin{axis}[
xmode=log,
width=1.2\figh,
height=.7\figh,
scale only axis,
xmin=1, xmax=1000,
xlabel={Regularization parameter $\nu$ ($\log$)},
ymin=0, ymax=4000,
ymajorgrids,
legend style={at={(0.03,0.03)},anchor=south west,draw=black,fill=white}]
\addplot [
color=green!60!black,
solid,
line width=1.25pt,
mark size=1.4pt,
mark=*,
mark options={solid,fill=green!60!black}
]
coordinates{(    1,   404) ( 3.16,   561) (   10,   986) (31.62,  1562) (  100,  2175) (316.22,  3051) ( 1000,  3402)};

\addplot [
color=green!60!black,
dashed,
line width=1.25pt,
mark size=0pt,
mark=*,
mark options={solid,fill=blue}
]
coordinates{(    1, 3726) ( 1000, 3726)};

\end{axis}
\end{tikzpicture}%

Tree depth $d$ =16
\end{minipage}

%% file: figs/timing/timing.tex
\begin{wrapfigure}{r}{5.3cm}
  \centering
  \vspace{-.55cm}
  \includegraphics[width=4.5cm]{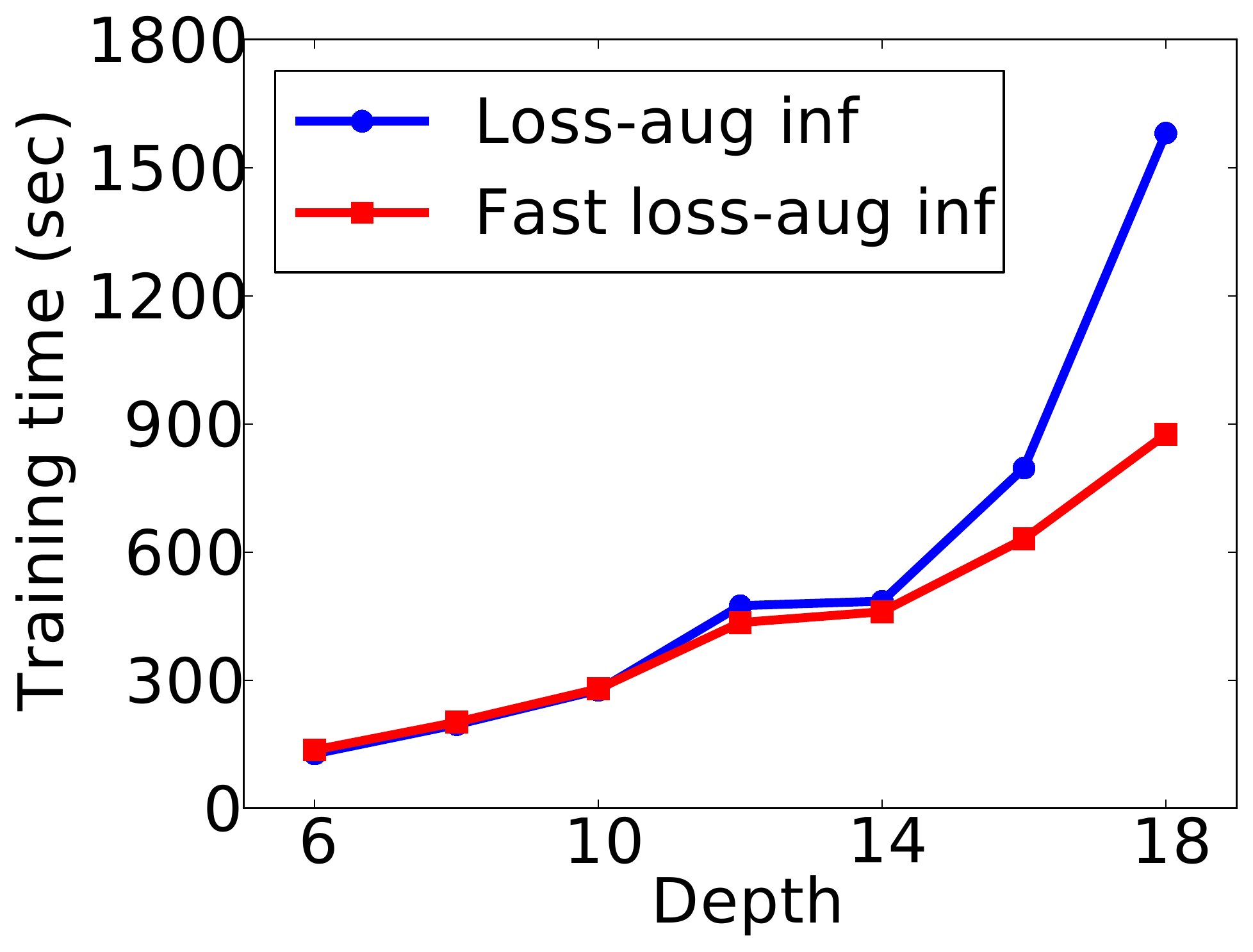}
  \vspace{-.2cm}
  \caption{ Total time to execute $1000$ epochs of SGD on the Connect4
    dataset using loss-agumented inference and its fast varient. }
  \label{fig:timing}
  \vspace{-.3cm}
\end{wrapfigure}

%% file: discussion.tex
\vspace*{-.1cm}
\section{Conclusion}
\vspace*{-.1cm}


We present a non-greedy method for learning decision trees using
stochastic gradient descent to optimize an upper bound on the tree's
empirical loss on a training dataset.  Our model poses the global
training of decision trees in a well-characterized optimization
framework.  This makes it simpler to pose extensions that could be
considered in future work.  Efficiency gains could be achieved by
learning sparse split functions via sparsity-inducing regularization
on $\W$.  Further, the core optimization problem permits applying the
kernel trick to the linear split parameters $\W$, making our overall
model applicable to learning higher-order split functions or training
decision trees on examples in arbitrary reproducing kernel Hilbert
spaces.

{\bf Acknowledgment.} MN was financially supported in part by a Google
fellowship. DF was financially supported in part by NSERC Canada and
the NCAP program of the CIFAR.

%% file: appendix.tex

\section{Proofs}

\setcounter{theorem}{0}

{\bf Upper bound on loss.} For any pair $(\x, \y)$, the loss $\loss(\trans{\Leaves} f(\sign(\W \x)), \y)$ is upper bounded by: 
\begin{equation}
\loss(\trans{\Leaves} f(\sign(\W \x)), \y) ~\le~  \max_{\g \in \Hset} \big\{ \, \trans{\g}\W\x + \loss(\trans{\Leaves} f(\g), \y) \big\} - \, \max_{\h \in \Hset} \,\big\{ \trans{\h}\W\x \big\}~.
\end{equation}

\begin{proof}
\begin{eqnarray*}
\mbox{RHS} &=& \max_{\g \in \Hset} \big\{ \, \trans{\g}\W\x + \loss(\trans{\Leaves} f(\g), \y) \big\}
- \, \max_{\h \in \Hset} \,\big\{ \trans{\h}\W\x \big\}\\
&=& \max_{\g \in \Hset} \big\{ \, \trans{\g}\W\x + \loss(\trans{\Leaves} f(\g), \y) \big\}
- \, \sign (\W\x)^T \W\x \\
&\geq& \max_{\g \in \{\sign (\W\x)\}} \big\{ \, \trans{\g}\W\x + \loss(\trans{\Leaves} f(\g), \y) \big\}
- \, \sign (\W\x)^T \W\x \\
&=& \sign (\W\x)^T \W\x + \loss(\trans{\Leaves} f(\sign (\W\x)), \y)
- \, \sign (\W\x)^T \W\x \\
&=& \loss(\trans{\Leaves} f(\sign(\W \x)), \y) \\
&=& \mbox{LHS}
\end{eqnarray*}
\end{proof}

\begin{proposition}
The upper bound on the loss becomes tighter as a constant multiple of
$\W$ gets larger. More formally, for any $\alpha > \beta > 0$, we have:
\begin{eqnarray}
&&\hspace{-.5cm}\max_{\g \in \Hset} \big\{ \alpha\trans{\g}\W\x + \loss(\trans{\Leaves} f(\g), \y) \big\} -  \max_{\h \in \Hset} \big\{ \alpha\trans{\h}\W\x \big\} ~\le~ \nonumber\\ &&\hspace{3cm}\max_{\g' \in \Hset} \big\{  \beta\trans{\g'}\W\x + \loss(\trans{\Leaves} f(\g'), \y) \big\} -  \max_{\h' \in \Hset} \big\{ \beta\trans{\h'}\W\x \big\}~.
\end{eqnarray}
\begin{proof}
Let
\begin{equation*}
\widehat{\g}_\alpha = \argmax{\g \in \Hset} \big\{ \alpha\trans{\g}\W\x + \loss(\trans{\Leaves} f(\g), \y) \big\}~,~~~~~~~~\widehat{\g}_\beta = \argmax{\g \in \Hset} \big\{ \beta\ \trans{\g}\W\x + \loss(\trans{\Leaves} f(\g), \y) \big\}~,
\end{equation*}
then we have:
\begin{equation}
\beta\ \trans{\widehat{\g}_\alpha}\W\x + \loss(\trans{\Leaves} f(\widehat{\g}_\alpha), \y) ~~\le~~ \beta\ \trans{\widehat{\g}_\beta}\W\x + \loss(\trans{\Leaves} f(\widehat{\g}_\beta), \y)~.
\label{eq:1st}
\end{equation}
We also have:
\begin{equation}
\max_{\h \in \Hset} \big\{ \alpha\ \trans{\h}\W\x \big\} ~=~ \alpha \ \trans{\sign(\W\x)}\W\x~,~~~~~~~\text{and}~~~~~~~\max_{\h \in \Hset} \big\{ \beta\ \trans{\h}\W\x \big\} ~=~ \beta \ \trans{\sign(\W\x)}\W\x~.
\label{eq:sgn-alpha-beta}
\end{equation}
Moreover,
\begin{eqnarray}
\trans{\widehat{\g}_\alpha}\W\x &\le& \trans{\sign(\W\x)}\W\x~~\implies\nonumber\\
(\alpha-\beta)\ \trans{\widehat{\g}_\alpha}\W\x &\le& (\alpha-\beta)\ \trans{\sign(\W\x)}\W\x~~\implies\nonumber\\
(\alpha-\beta)\ \trans{\widehat{\g}_\alpha}\W\x -\alpha\ \trans{\sign(\W\x)}\W\x &\le& -\beta\ \trans{\sign(\W\x)}\W\x~.
\label{eq:2nd}
\end{eqnarray}
Now, summing the two sides of \eqref{eq:1st} and \eqref{eq:2nd}, and using \eqref{eq:sgn-alpha-beta}, the inequality is proved.
\end{proof}
\end{proposition}